\pdfoutput=1
\documentclass[11pt]{article}

\usepackage[preprint]{acl}

\usepackage{times}
\usepackage{latexsym}

\usepackage[T1]{fontenc}
\usepackage[utf8]{inputenc}

\usepackage{microtype}

\usepackage{inconsolata}

\usepackage{graphicx}

\usepackage{amsmath}

\title{Hidden Persuasion: Detecting Manipulative Narratives on Social Media During the 2022 Russian Invasion of Ukraine}

\author{Kateryna Akhynko \\
  Ukrainian Catholic University \\
  Lviv, Ukraine \\
  \texttt{kateryna.akhynko@ucu.edu.ua} \\\And
  Oleksandr Kosovan \\
  Ukrainian Catholic University \\
  Lviv, Ukraine \\
  \texttt{o.kosovan@ucu.edu.ua} \\\And
  Mykola Trokhymovych \\
  Pompeu Fabra University \\
  Barcelona, Spain \\
  \texttt{mykola.trokhymovych@upf.edu} \\}

\begin{document}
\maketitle

\begingroup
\renewcommand\thefootnote{}\footnote{To appear in UNLP’25.}
\addtocounter{footnote}{-1}
\endgroup

\begin{abstract}
This paper presents one of the top-performing solutions to the UNLP 2025 Shared Task on Detecting Manipulation in Social Media. The task focuses on detecting and classifying rhetorical and stylistic manipulation techniques used to influence Ukrainian Telegram users. For the classification subtask, we fine-tuned the Gemma 2 language model with LoRA adapters and applied a second-level classifier leveraging meta-features and threshold optimization. For span detection, we employed an XLM-RoBERTa model trained for multi-target, including token binary classification. Our approach achieved 2nd place in classification and 3rd place in span detection. 
\end{abstract}

\section{Introduction}

In times of war, information can have the same power as weaponry. During the 2022 Russian invasion of Ukraine, Telegram emerged not only as a battlefield communication tool but also as the primary source of information for 44\% of Ukrainians. Its speed, reach, and anonymity became an important tool for civilians and military actors. However, these features — particularly minimal content moderation and user anonymity — have also made Telegram a favorable environment for influence operations~\cite{vorobiov2024telegram}.

Manipulation on social media is a complex and nuanced phenomenon. It includes not just factual distortions (i.e., disinformation) but also rhetorical strategies, emotional appeals, and narrative framing that are designed to influence perception or behavior subtly. 
In this paper, we present the solution\footnote{\url{https://github.com/akhynkokateryna/manipulative-narrative-detection}} to the UNLP 2025 Shared Task,\footnote{\url{https://github.com/unlp-workshop/unlp-2025-shared-task}} focused on manipulative narratives detection, which is defined as the intentional use of language and messaging tactics aimed at influencing beliefs, emotions, or attitudes, without providing clear factual support.

% Detecting such manipulations forms a unique set of challenges. 
The task includes several challenges that make it particularly complex.
First, it focuses exclusively on the textual content of social media posts without incorporating metadata such as user history or engagement metrics. Second, the dataset presents multiple layers of complexity: it is imbalanced across manipulation types, multilingual (primarily Ukrainian and Russian), and multi-label, meaning that a single post can include several manipulation techniques simultaneously. Finally, the span detection subtask requires identifying the exact textual fragments responsible for the manipulation, often implicit, rhetorical, or emotionally charged language that is difficult to isolate.

\begin{figure}[t]
  \includegraphics[width=\columnwidth]{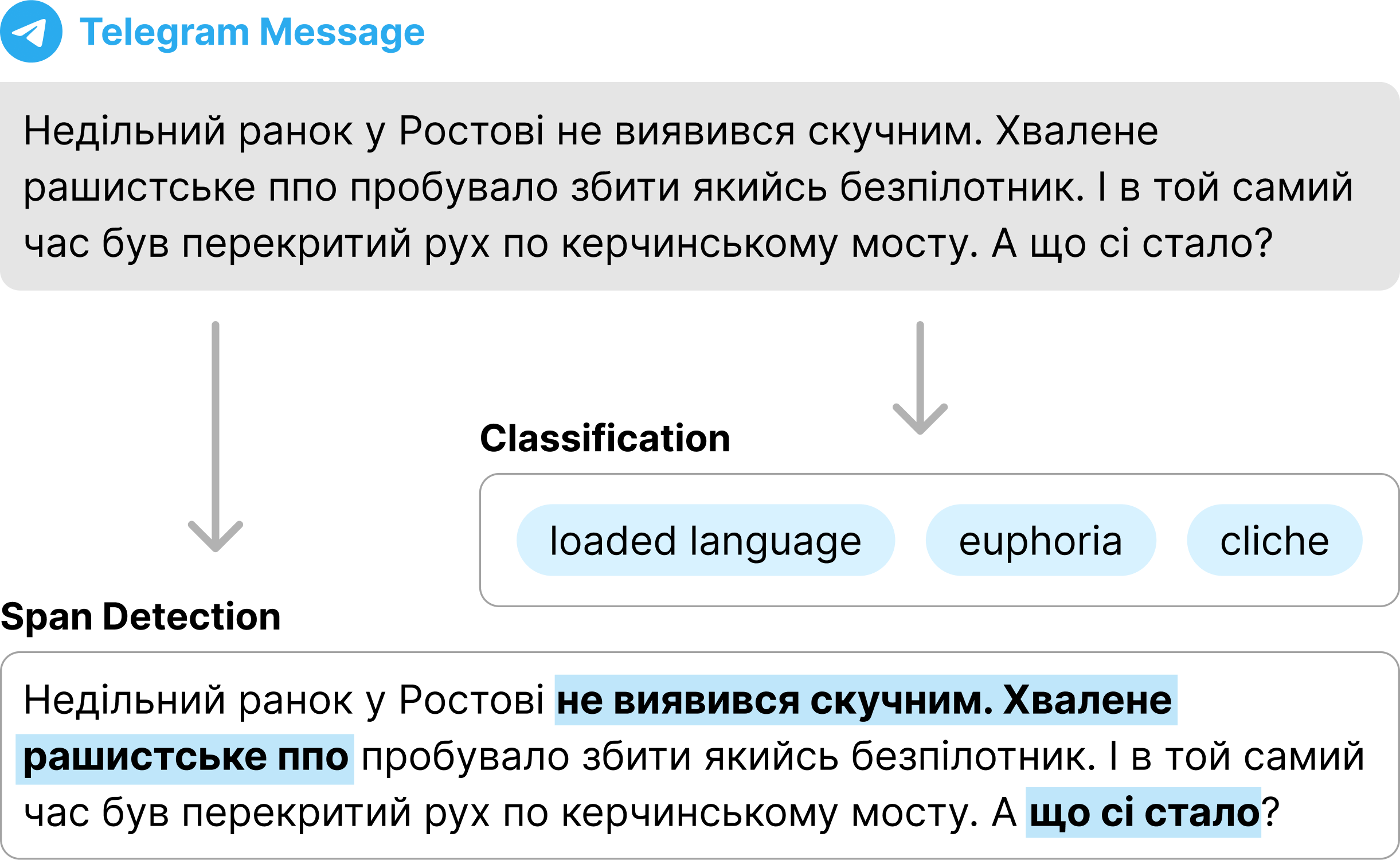}
  \caption{Sketch of manipulation techniques classification and span detection problems}
  \label{fig:sketch}
\end{figure}

Given these challenges, we developed a system that achieved second place in manipulation techniques classification and third place in span detection subtasks (see Figure~\ref{fig:sketch}). For classification, we fine-tuned the Gemma 2 language model using LoRA adapters and introduced a second-level classifier that leveraged meta-features and custom threshold optimization. For span detection, we trained an XLM-RoBERTa model capable of multi-target, token-level binary classification to locate manipulative spans within posts.

\section{Related Work}

Our research is based on a growing body of work in detecting propaganda and misinformation analysis. Numerous studies have focused on identifying propaganda techniques in news articles, particularly in the context of SemEval-2020 Task 11. \citet{da-san-martino-etal-2020-semeval} explored detecting propaganda techniques in news articles through span identification and technique classification tasks. 

Similarly to previous research, the UNLP 2025 Shared Task includes two subtasks: manipulation technique classification (a multi-label classification) and span detection (a token classification). Within this framework, research from SemEval-2020 Task 11 demonstrated BERT's remarkable capabilities for propaganda technique identification \cite{altiti-etal-2020-just}. Further advancing this line of inquiry, \citet{da-san-martino-etal-2020-semeval} showcased RoBERTa's performance in addressing both tasks simultaneously.

% At the same time, the nature of propaganda in social media changes over time and varies with the specific circumstances, adapting to remain undetected. \citet{solopova-etal-2023-evolution} examined this process, blending insights from machine learning and linguistics to reveal how pro-Kremlin propaganda evolves in the context of the 2022 Russian invasion of Ukraine. At the same time, in our task, we concentrate on detecting manipulative narratives without importance of if text is factually correct, that makes our task different from the knowledge manipulation or fact-checking. 
At the same time, the nature of propaganda on social media evolves continuously, adapting to specific circumstances to remain undetected. \citet{solopova-etal-2023-evolution} explored this process by combining machine learning and linguistic analysis to reveal how pro-Kremlin propaganda evolved in the context of the 2022 Russian invasion of Ukraine. In this context, it is important to note that while our work has a similar goal, we focus specifically on detecting manipulative narratives regardless of the factual support of the claim. This distinguishes our approach from fact-checking or knowledge manipulation detection methods~\cite{10.1145/3459637.3481961,trokhymovych2025}.

In our case, we are dealing with multilingual Telegram data containing Ukrainian and Russian texts. In this scenario, fine-tuning a multilingual model, such as XLM-RoBERTa, appears to be a more productive approach, as demonstrated in research on hostility identification for low-resource Indian languages \cite{hostility-identification}. Moreover, XLM-RoBERTa-based models have demonstrated cross-lingual strengths in other downstream tasks, including those involving Ukrainian and Russian languages~\cite{mehta-varma-2023-llm,trokhymovych-etal-2024-open}.

While \citet{sprenkamp2023largelanguagemodelspropaganda} discovered that fine-tuned RoBERTa outperformed zero and few-shot learning approaches with LLMs for propaganda detection, newer advances in large language models show considerable promise. Recent innovations have developed methods to transform decoder-only LLMs into effective text encoders suitable for classification tasks \cite{behnamghader2024llm2veclargelanguagemodels}. Models such as Gemma offer particularly interesting customization potential for classification challenges \cite{gemmateam2024gemma2improvingopen}.

Notably, Gemma-family models enable fine-tuning with LoRA adapters and support quantization techniques, making them viable options even with limited computational resources. Building on this foundation, \cite{kiulian2024bytesborschfinetuninggemma} ventured into fine-tuning both Gemma and Mistral specifically to enhance Ukrainian language representation, providing valuable insights that directly inform our approach to detecting manipulative narratives within Telegram content from the region.

\section{Data}

The UNLP shared task dataset contains more than 9,500 text samples collected from Telegram channels, with 68\% of these collected samples containing manipulative narratives. 
This dataset forms the basis for a dual-task challenge: classifying manipulation techniques and identifying corresponding text spans.

The data is divided into training and testing sets, with 3,822 samples allocated for training and 5,735 for testing. 
Among the 3,822 training samples, 2,147 (56\%) are in Ukrainian and 1,675 (44\%) are in Russian.
At the same time, the testing set does not include language labels.
Notably, the testing set is further split into public and private sets for leaderboard evaluation. 
% It is important to emphasize that the split is not based on temporal order. As a result, older posts may appear in the testing set, potentially limiting the real-world applicability of models trained on this data due to the lack of chronological separation.

\begin{figure}[t]
  \includegraphics[width=\columnwidth]{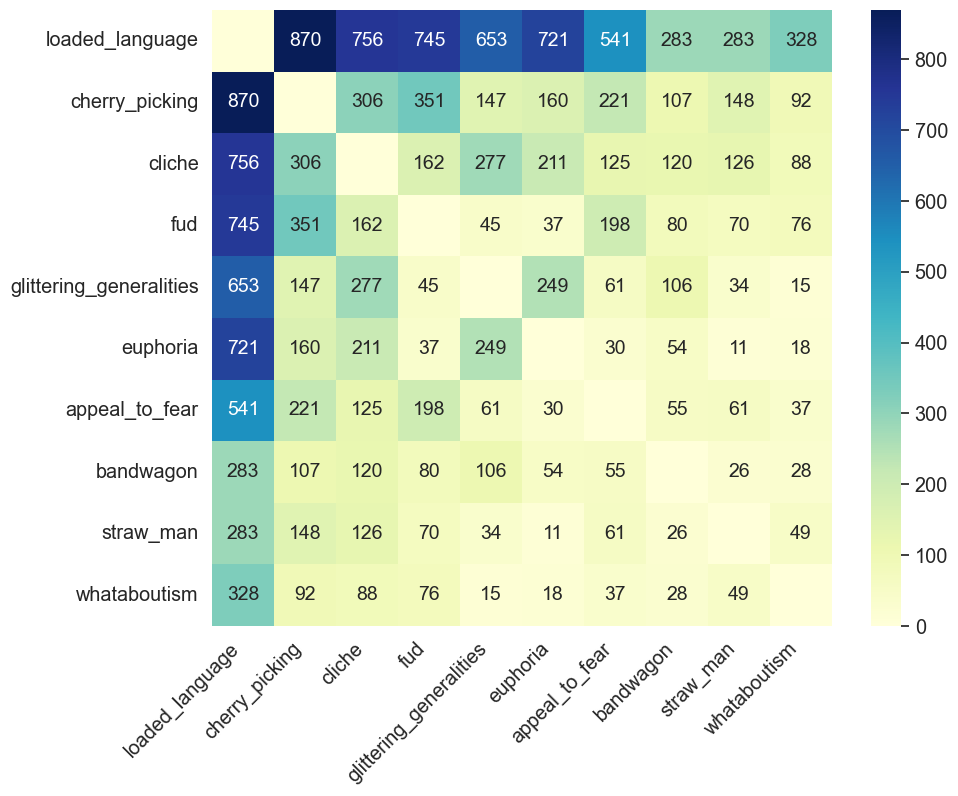}
  \caption{Co-occurrence of manipulation techniques in the combined training and testing sets}
  \label{fig:co-occurrence}
\end{figure}

Each post is annotated for both classification and span detection tasks. Specifically, every sample is labeled with one or more of ten predefined manipulation techniques, detailed in Appendix~\ref{sec:manipulation-techniques}. Manipulative text segments are also defined, irrespective of the specific technique involved.

Figure~\ref{fig:co-occurrence} illustrates manipulation techniques' co-occurrence patterns across training and testing sets. As the distribution of labels is similar in both subsets, we present them together for clarity.

\section{Methodology}

In this section, we present our approaches for solving the technique classification and span identification subtasks.

\subsection{Technique Classification}
\label{sec:tech_class}

The manipulation technique detection task is formulated as a multi-label text classification problem, where each input text may contain multiple manipulation strategies. Each sample is annotated with any number of 10 predefined manipulation techniques.

\begin{figure*}[t]
  \centering
  \includegraphics[width=\linewidth]{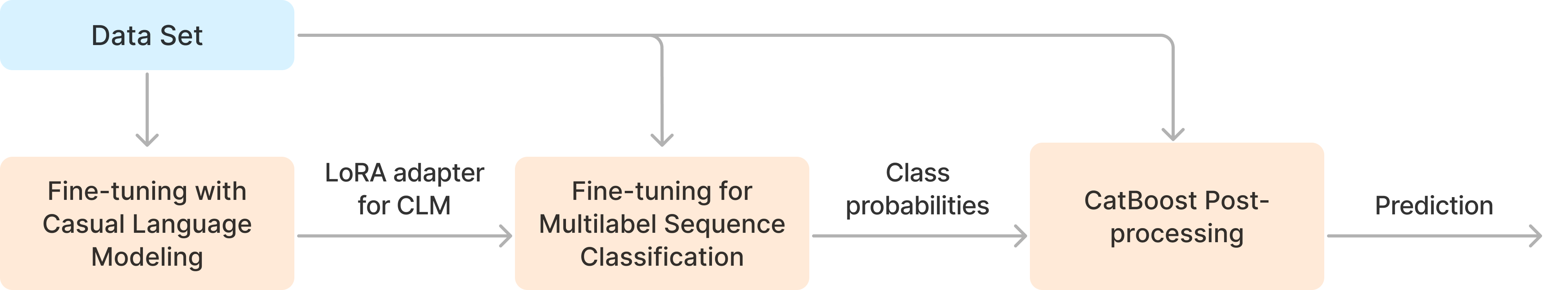}
  \caption{Pipeline of technique classification solution}
  \label{fig:technique-classification}
\end{figure*}

Our best-performing solution involves multi-stage fine-tuning of the instruction-tuned Gemma 2 2B IT model.\footnote{\url{https://huggingface.co/google/gemma-2-2b-it}} The complete fine-tuning pipeline schema is presented in Figure~\ref{fig:technique-classification}. 

\textbf{Firstly,} we fine-tune the model using a causal language modeling (CLM) objective, where the model learns to predict the next token given a left-to-right context. Specifically, we employed the \texttt{AutoModelForCausalLM} class from \texttt{HuggingFace Transformers}.

The model was trained to autoregressively generate a comma-separated list of manipulation techniques based on a task-specific prompt. We constructed a dataset of prompt inputs for each training data point, which included:
\begin{itemize}
    \item an instruction to identify manipulative techniques in a text;
    \item descriptions of all ten manipulation techniques;
    \item four few-shot examples, selected from the training set: two were chosen based on cosine similarity between the target text and other texts in the training set, and the other two based on cosine similarity between the target text and the trigger phrases (i.e., manipulative spans in texts) found in other training samples.
\end{itemize}

% To control input length, we limited selected texts to a maximum of 500 characters. 
To control input length, we select the few-shot examples from the subset limited by texts shorter than 500 characters. 
To get a vector representation of the texts, we encode them using SentenceTransformers, employing \textit{mGTE} model~\cite{reimers-2019-sentence-bert,zhang2024mgte}. Later, these vectors are used for few-shot candidates selection and text clustering. 

As for this stage of model tuning, we used almost the whole training dataset, as our main goal was to expose the model to as much relevant data as possible rather than tuning to a specific downstream task.
Due to the high computational cost of full model fine-tuning, we instead trained LoRA adapter using a CLM objective. The adapter was configured with causal LM task type via the PEFT library to ensure compatibility with the CLM setup.
% The adapter was configured to update only a minimal set of attention-related layers, following common practice in language modeling. 
% To further speed up training, we applied quantization with standard settings. 
Finally, we got the fine-tuned adapter for the text generation in the form of a list of manipulation techniques. 
% This approach adapts the model's internal representation to our task, particularly the last sequence token, which is crucial for later multi-label classification.

\textbf{In the second stage,} we merged the LoRA adapter from the first stage with the base model, set the model to a multi-label classification mode, and trained an additional LoRA adapter.
The input for this stage consisted of text samples and their corresponding technique labels.

\textbf{In the third stage,} we combined the probability outputs from the previous stage with a set of engineered meta-features to train a CatBoost model for multi-label classification on the same training set. The additional features include:

\begin{enumerate}
    \item distances from each text to the centroids of clusters formed by triggered phrases from the training set using K-means;
    \item frequency of each manipulation technique among the most similar examples from the training set selected based on cosine similarity with their text and trigger phrases;
    \item additional meta-features such as word count, number of question marks, presence of URLs, etc.
\end{enumerate}

To construct the clustering-based features, we applied the K-means clustering algorithm to the set of triggered phrases extracted from the training set. Firstly, we encode the text with SentenceTransformers as mentioned earlier. 
% Before clustering, the phrases were encoded using SentenceTransformers, employing \textit{mGTE} model~\cite{reimers-2019-sentence-bert,zhang2024mgte}. 
We set the number of clusters (K) to be K=10, equal to the number of unique manipulation techniques. Finally, for each sample text, we calculate the cosine distance to the centroid of each cluster. This approach allows the model to capture how semantically close a text is to common manipulation patterns identified in the training data.

For the similarity-based frequency features, we computed pairwise cosine similarity between the embedded texts. For each text, we selected two sets of 10 most similar examples from the training set: (1) based on overall similarity to other full texts, and (2) based on similarity to trigger phrases from other texts. We calculated the frequency distribution of manipulation techniques among the nearest neighbors in both cases.
These techniques and meta-linguistic features (e.g., word count, presence of punctuation) were combined with model probabilities to train the final CatBoost classifier.

Finally, since the dataset is highly imbalanced, we optimized class-wise thresholds by performing k-fold cross-validation and choosing the median of the best thresholds within folds for each class separately. This approach avoids the pitfalls of using a single global threshold, especially for rare classes, and improves overall performance on the macro F1 score, which treats all classes equally. So, we used this method to construct the final prediction using the probability scores from the CatBoost model.

\subsection{Span Identification}
\label{sec:span-method}

Span identification for manipulative content is defined as a binary token classification task, where each token is labeled as either manipulative or non-manipulative, independent of the specific manipulation technique. Identified manipulative tokens are then mapped to character indices and grouped into spans, allowing for precise extraction of manipulative text.

\begin{figure}[t]
  \includegraphics[width=\columnwidth]{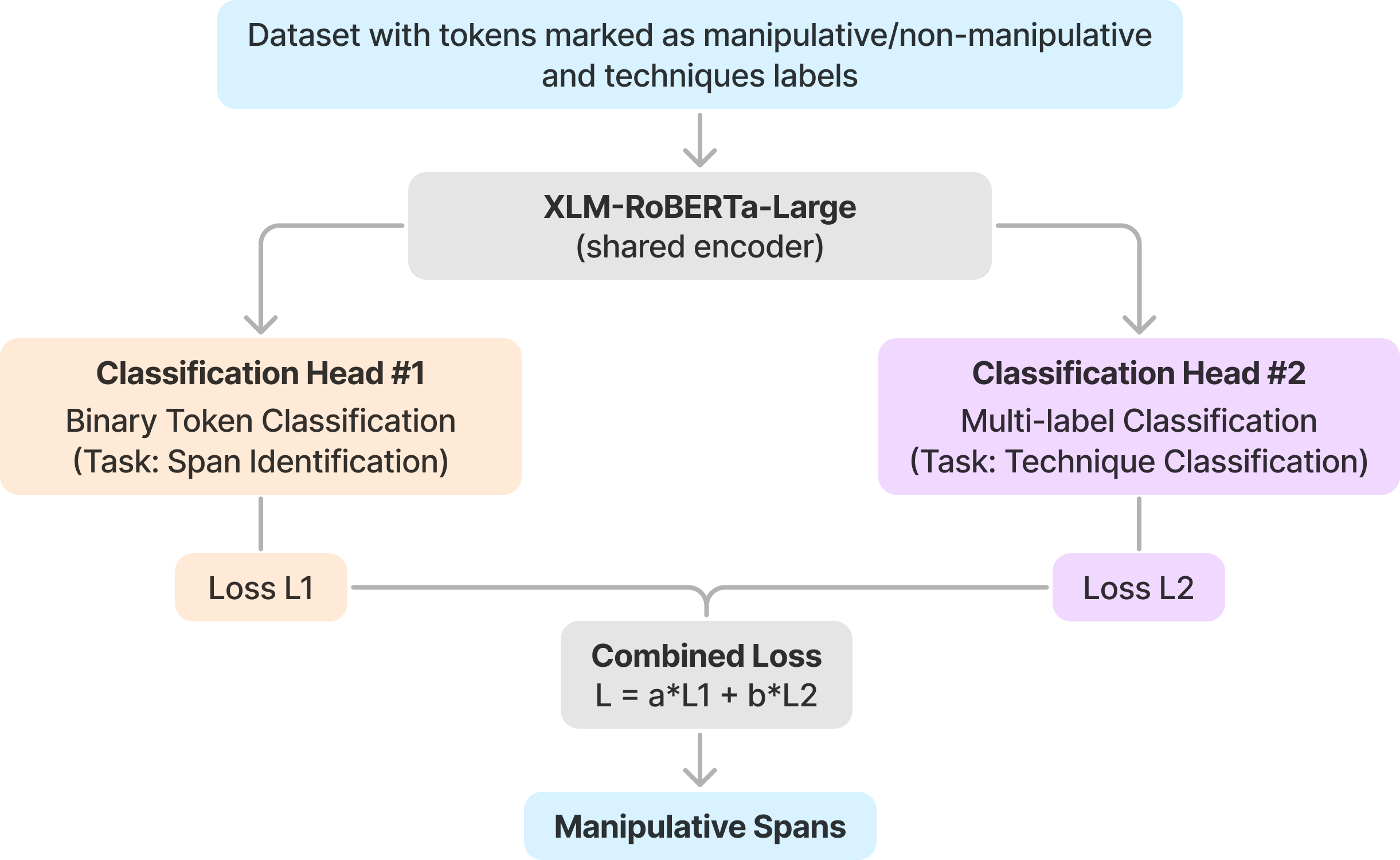}
  \caption{Pipeline of span identification solution}
  \label{fig:two-head-model}
\end{figure}

For this task, we employ a multi-headed architecture based on the XLM-RoBERTa-Large\footnote{\url{https://huggingface.co/FacebookAI/xlm-roberta-large}} (see Figure~\ref{fig:two-head-model}). Two custom classification heads are introduced: one dedicated to classifying manipulative techniques (multi-label classification) and the other to token classification. Both heads share a common encoder, allowing the model to benefit from shared representations across tasks.

The span identification head consists of a single linear layer applied to the contextualized token representations, predicting the likelihood of each token being part of a manipulative span.

The technique classification head operates on a pooled representation formed by concatenating the \textit{[CLS]} token embedding, mean-pooled, and max-pooled token embeddings. This concatenated vector is passed through a linear layer that projects it to a lower-dimensional space of size 256, followed by a GELU activation. The intermediate representation is then regularized through layer normalization and dropout before being passed to a final linear layer that projects it to the space of manipulation technique labels.

% This shared encoding enables knowledge transfer between the two subtasks—semantic understanding developed for technique classification can inform the model’s span detection decisions, particularly in our low-resource case. 

To balance the influence of both tasks during training, we apply a reduced weighting coefficient to the classification head’s loss when computing the overall objective. This ensures that span detection remains the primary focus, while the model still benefits from auxiliary guidance.

Consistent with Technique Classification Subtask, we determine optimal prediction thresholds through k-fold cross-validation, ensuring robust calibration and generalization across splits.

\section{Evaluation}

\subsection{Technique Classification}

The manipulation techniques classification subtask, as defined in the shared task, uses a macro-averaged F1 score as its primary evaluation metric. This metric treats all classes equally, regardless of their frequency in the dataset. Appendix~\ref{sec:techniques-metric} provides a detailed explanation of the metric.

% Before submitting our final inference for leaderboard evaluation, we developed and relied on a local evaluation pipeline. For multi-stage solutions, we assessed performance at each stage individually to ensure consistency and reliability. The approach illustrated in Fig.~\ref{fig:technique-classification} involves three distinct stages of training and evaluation. The first stage is a Causal Language Modeling setup, where the dataset, augmented with prompts, is split into training and testing subsets. In the second stage, we perform sequence classification, again splitting the data into train/test sets and selecting appropriate thresholds. The final stage incorporates CatBoost-based postprocessing, using features and predicted scores from the previous step, followed by another train/test split and threshold optimization.

% To handle class imbalance, we applied per-class threshold optimization with 5-fold stratified cross-validation. For each label, we selected thresholds that maximized the F1 score in validation folds, repeating the process independently per class. Median values across folds were then used as final thresholds. This approach led to a more balanced evaluation, improving performance on underrepresented classes compared to using a fixed global threshold.

Our main results are summarized in Table~\ref{tab:classification-metrics}, where F1 scores were recalculated on the full testing set. As a baseline, we used a multi-label CatBoost model with threshold optimization. For baseline training, we use a dataset that consists only of meta-features used in the final step, as explained in Section~\ref{sec:tech_class}.

Although the baseline appeared to be an effective solution regarding resource efficiency and performance, it was insufficient to remain competitive in the challenge. This motivated integrating the Gemma 2-based solution, as introduced in Section~\ref{sec:tech_class}. In our final comparison, we present two configurations of this model—with and without final post-processing using CatBoost and metafeatures. The results demonstrate that Gemma-based solutions significantly outperform the baseline. 
Although the post-processing step results in only a minor improvement, it is essential to achieve a competitive advantage in the competition.
% Although the post-processing step results in only a minor improvement, achieving a competitive advantage in the competition is essential.

% The next major step was the integration of the Gemma model, which led to a significant performance boost. Finally, we applied post-processing, resulting in small score improvements. This solution achieved second place in the competition, as shown in the Table \ref{tab:classification-metrics-leader-board}. 

We also conducted a performance analysis for each class (see Table~\ref{tab:classification-report}), revealing considerable variation in the model’s effectiveness across different techniques. Notably, the model performs significantly worse on underrepresented classes such as \textit{whataboutism}, \textit{straw\_man}, and \textit{bandwagon}. In contrast, it achieves the highest performance on the \textit{loaded\_language} class, which has over ten times more samples than the mentioned underrepresented ones. 

% Additionally, we made an error analysis of the final solution inference (Fig. \ref{fig:confusion-heatmap}), which showed us what mistakes our model made most often (for example, we should pay more attention to the loaded language class). Overall, these results suggest that the Gemma model is a promising approach for this classification task.

% \begin{figure}[t]
%   \includegraphics[width=\columnwidth]{latex/img/multilabel_pred_heatmap_norm.png}
%   \caption{Normalized Confusion Heatmap for Multilabel Classification}
%   \label{fig:confusion-heatmap}
% \end{figure}

\begin{table}[tb]
    \centering
    \begin{tabular}{lcccc}
        \hline \textbf{Technique}
         & \textbf{F1 score} & \textbf{Support} \\
        \hline
        \textbf{appeal\_to\_fear} & 0.450 & 449 \\
        \textbf{bandwagon} & 0.215 & 236 \\
        \textbf{cherry\_picking} & 0.467 & 768 \\
        \textbf{cliche} & 0.328 & 695 \\
        \textbf{euphoria} & 0.550 & 695 \\
        \textbf{fud} & 0.525 & 576 \\
        \textbf{glittering\_generalities} & 0.644 & 723 \\
        \textbf{loaded\_language} & 0.782 & 2959 \\
        \textbf{straw\_man} & 0.287 & 207 \\
        \textbf{whataboutism} & 0.296 & 235 \\
        % \hline
        % \textbf{micro avg} & 0.572 & 7543 \\
        % \textbf{macro avg} & 0.454 & 7543 \\
        % \textbf{weighted avg} & 0.588 & 7543 \\
        % \textbf{samples avg} & 0.375 & 7543 \\
        \hline
    \end{tabular}
    \caption{Classification report for technique prediction}
    \label{tab:classification-report}
\end{table}

\begin{table}[tb]
  \centering
  \begin{tabular}{lcc}
    \hline
    \textbf{Solution} & \textbf{F1 macro} \\
    \hline
    Baseline (CatBoost)     & 0.40801 \\
    Gemma                  & 0.45007 \\
    Gemma with post-processing  & 0.45447 \\
    \hline
  \end{tabular}
  \caption{Comparison of our solutions for technique classification during the competition}
  \label{tab:classification-metrics}
\end{table}

\begin{table}
  \centering
  \begin{tabular}{lcc}
    \hline
    \textbf{Team} & \textbf{Public} & \textbf{Private} \\
    \hline
    GA                          & 0.47369            & 0.49439            \\
    \textbf{MolodiAmbitni}      & \textbf{0.46203}   & \textbf{0.46952}   \\
    CVisBetter\_SEU             & 0.43669            & 0.45519            \\
    \hline
  \end{tabular}
  % \caption{Classification metrics from leaderboard}
\caption{Comparison of metrics for top-3 solutions from competition leaderboard for manipulation classification}
  \label{tab:classification-metrics-leader-board}
\end{table}

\subsection{Span Identification}

Like the previous subtask, span identification relies on the evaluation metrics defined in the shared task. Specifically, we use the span-level F1-score, quantifying the overlap between predicted and defined character spans. Appendix~\ref{sec:span-metric} provides a detailed explanation of this metric.

Our span detection pipeline also incorporates post-processing and a threshold selection step, as described in Section~\ref{sec:span-method}. As a strong baseline, we employed the XLM-RoBERTa model configured for token classification. Building on top of it, we explored the hypothesis that a two-head transformer, combined to address both subtasks simultaneously, could enhance generalization and improve results. Although, as shown in Table~\ref{tab:span-metrics}, the performance gain was not large. This approach ultimately secured us third place in the competition, as reported in Table~\ref{tab:span-metrics-leader-board}. These findings suggest that, for practical applications, a simpler baseline approach may be more robust and justified.

\begin{table}[t]
  \centering
  \begin{tabular}{lcc}
    \hline
    \textbf{Solution} & \textbf{Span-level F1} \\
    \hline
    Baseline    &   0.58588           \\
    Two-head transformer     & 0.59888          \\
    \hline
  \end{tabular}
  \caption{Comparison of our solutions for span detection during the competition}
  \label{tab:span-metrics}
\end{table}

\begin{table}[t]
  \centering
  \begin{tabular}{lcc}
    \hline
    \textbf{Solution} & \textbf{Public} & \textbf{Private} \\
    \hline
    GA                  & 0.64598 & 0.64058     \\
    CVisBetter\_SEU     & 0.59873 & 0.60456     \\
    \textbf{MolodiAmbitni}       & \textbf{0.59662} & \textbf{0.60001}     \\
    \hline
  \end{tabular}
  \caption{Comparison of metrics for top-3 solutions from competition leaderboard for span detection subtask}
  \label{tab:span-metrics-leader-board}
\end{table}

\section{Conclusion}

To sum up, this paper presents a competitive solution to the UNLP 2025 Shared Task on detecting manipulative narratives in Ukrainian Telegram news. By leveraging a multi-stage fine-tuned Gemma 2 language model with LoRA adapters for technique classification and a two-headed XLM-RoBERTa architecture for span detection, our approach secured second and third place in the respective subtasks.

Key achievements include a two-phase fine-tuning of a decoder-only model (Gemma) for classification, first via causal language modeling, then supervised multi-label learning. We further enhanced performance with a post-processing step using a CatBoost classifier that combined meta-features with previously predicted class probabilities. Per-class threshold optimization addressed label imbalance and improved macro-F1 performance.
For span detection, we introduced a dual-head architecture that jointly learned classification and token-level labeling, encouraging better generalization through shared representations.

Results show that each enhancement added measurable value. Post-processing raised the classification macro-F1 from 0.45007 to 0.45447, while span detection improved from 0.58588 to 0.59888 with the dual-head setup. However, performance varied notably across manipulation types: while frequent classes like \textit{loaded\_language} were predicted with high accuracy, rarer classes such as \textit{whataboutism} and \textit{straw\_man} remained challenging.

\section*{Limitations} 

We are working with a dataset that includes texts only in Ukrainian and Russian. While LLMs are improving multilingual support, existing open-source models have limited support for those languages. Also, Telegram posts often contain informal language, slang, neologisms, emojis, and irregular formatting. It may reduce the effectiveness of pre-trained models, which are typically trained on more formal text. 

While the dataset was annotated by experienced professionals, the manipulation signal is subjective and context-dependent. This can lead to ambiguous labels, especially in span identification, where the boundaries of manipulative content are not always clearly defined. 

Moreover, the dominance of certain manipulation techniques (e.g., loaded language) makes the classification task imbalanced.
% which can affect the ability of models to learn less frequent techniques. 
Although steps can be taken to mitigate this (e.g., resampling, class weighting, or threshold selection in our case), performance on rare techniques remains a challenge.

The dataset presented for the competition appears to be 
% randomly 
divided into training and test sets without considering the chronological order of posts. As a result, the evaluation may not reflect the real-world scenario of predicting new, emerging manipulation patterns. 
% This setup limits our ability to assess how well models generalize to future manipulative content, which is particularly important in dynamic information environments such as social media.

\section*{Acknowledgments}
We thank the Applied Sciences Faculty at Ukrainian Catholic University for providing access to computational resources that supported this research.

\bibliography{custom}

\appendix

\section{Manipulation Techniques}
\label{sec:manipulation-techniques}

Table \ref{technique-description} contains each class explanation that was provided by the organisers.\footnote{\url{https://github.com/unlp-workshop/unlp-2025-shared-task/blob/main/data/techniques-en.md}}

\begin{table*}
  \centering
  \resizebox{\textwidth}{!}{%
  \begin{tabular}{p{4cm} p{12cm}}
    \hline
    \textbf{Name}               & \textbf{Description} \\
    \hline
    Loaded Language             & The use of words and phrases with a strong emotional connotation (positive or negative) to influence the audience.  \\
    Glittering Generalities     & Exploitation of people's positive attitude towards abstract concepts such as “justice,” “freedom,” “democracy,” “patriotism,” “peace,” “happiness,” “love,” “truth,” “order,” etc. These words and phrases are intended to provoke strong emotional reactions and feelings of solidarity without providing specific information or arguments. \\
    Euphoria                    & Using an event that causes euphoria or a feeling of happiness, or a positive event to boost morale. This manipulation is often used to mobilize the population. \\
    Appeal to Fear              & The misuse of fear (often based on stereotypes or prejudices) to support a particular proposal. \\
    FUD (Fear, Uncertainty, Doubt) &Presenting information in a way that sows uncertainty and doubt, causing fear. This technique is a subtype of the appeal to fear. \\
    Bandwagon/Appeal to People  & An attempt to persuade the audience to join and take action because “others are doing the same thing.” \\
    Thought-Terminating Cliché  & Commonly used phrases that mitigate cognitive dissonance and block critical thinking. \\
    Whataboutism                & Discrediting the opponent's position by accusing them of hypocrisy without directly refuting their arguments. \\
    Cherry Picking              & Selective use of data or facts that support a hypothesis while ignoring counterarguments. \\
    Straw Man                   & Distorting the opponent's position by replacing it with a weaker or outwardly similar one and refuting it instead. \\
    \hline
  \end{tabular}
  }
  \caption{\label{technique-description}
    Explanation of Manipulation Techniques provided by UNLP Shared Task}
\end{table*}

\section{Metrics}

\subsection{Techniques Classification}
\label{sec:techniques-metric}

To evaluate the classification of manipulation techniques, we use the macro-averaged F1 score, which ensures balanced assessment across all techniques. Given a set of texts \( V \) and manipulation techniques \( T \), each text is labeled with a binary vector indicating the presence of techniques. The model predicts a vector of the same size, and for each technique \( t \in T \), we compute the F1 score:

\[
F1_t = \frac{2 \cdot P_t \cdot R_t}{P_t + R_t}
\]

where precision \( P_t \) measures correct predictions among all predicted instances, and recall \( R_t \) measures correct predictions among actual instances. The final macro-F1 score is obtained as:

\[
F1_{\text{macro}} = \frac{1}{|T|} \sum_{t \in T} F1_t
\]

This approach is particularly useful for handling class imbalances as it prevents frequently occurring techniques, which are typically detected with greater accuracy, from dominating the overall performance score.

\subsection{Span Identification}
\label{sec:span-metric}

To evaluate the accuracy of detected spans, we use the span-level F1 score, which measures the overlap between predicted and actual spans. Let \( V \) be the set of all texts in the dataset. Each text \( v \in V \) has a set of ground truth spans \( S_v \) and predicted spans \( \hat{S}_v \). The set of manipulated tokens in text \( v \) is defined as the collection of all characters whose index falls in at least one manipulation span:

\[
T_v = \bigcup_{(s, e) \in S_v} \{s, s+1, \dots, e-1\}
\]

\[
\hat{T}_v = \bigcup_{(s, e) \in \hat{S}_v} \{s, s+1, \dots, e-1\}
\]

Precision and recall are computed as:

\[
P = \frac{\sum_{v \in V} |T_v \cap \hat{T}_v|}{\sum_{v \in V} |\hat{T}_v|}
\]

\[
R = \frac{\sum_{v \in V} |T_v \cap \hat{T}_v|}{\sum_{v \in V} |T_v|}
\]

The final span-level F1 score is given by:

\[
F1 = \frac{2 P R}{P + R}
\]

% \section{Hyperparameters}

\end{document}